\documentclass[conference]{IEEEtran}
\usepackage{graphicx}
\usepackage{graphics,graphicx} 
\usepackage{epsfig} 
 \pdfoutput=1
\usepackage{multirow}
\usepackage{amsmath}
\usepackage{bm}

\usepackage{amssymb}
\usepackage{tikz}
\usepackage{amsthm}
\usepackage{comment}
\usepackage{mathcomSTEv4}
\usepackage{subcaption}
\usepackage{setspace}
\usepackage[utf8]{inputenc}
\usepackage{cite}

\usepackage{hyperref}
\hypersetup{
	colorlinks=true,
	linkcolor=blue,
	filecolor=magenta,      
	urlcolor=blue,
}

\usepackage{amsmath}
\usepackage{steinmetz}
\usepackage{color}
\usepackage{graphicx} 
\usepackage{epsfig} 

\usepackage{amsmath}
\usepackage{amssymb}
\usepackage{tikz}
\usepackage{amsthm}
\usepackage{comment}
\usepackage[export]{adjustbox}
\usepackage{algorithm}
\usepackage[noend]{algpseudocode}
\makeatletter
\def\BState{\State\hskip-\ALG@thistlm}
\makeatother

\newtheorem*{remark}{Remark}

\begin{document}

\title{
The Information \& Mutual Information Ratio\\ for Counting Image Features and Their Matches
}

\author{
	\IEEEauthorblockN{Ali Khajegili Mirabadi}
	\IEEEauthorblockA{
		Department of Electrical and Computer Engineering\\
		Isfahan University of Technology, Isfahan 84156-83111, Iran\\
		\texttt{ali.khajegili@ec.iut.ac.ir}
	}
	\and
	\IEEEauthorblockN{Stefano Rini}
	\IEEEauthorblockA{
		Department of Electrical and Computer Engineering\\
		National Chiao Tung University, Taiwan\\
		\texttt{stefano@nctu.edu.tw}
	}
}

\maketitle

\begin{abstract}
Feature extraction and description is an important topic of computer vision,  as it is the starting point of a number of tasks such as image reconstruction, stitching, registration, and recognition among many others. 
	In this paper, two new image features are proposed: the Information Ratio (IR) and the Mutual Information Ratio (MIR). 
	The IR is a feature of a single image, while the MIR describes features common across two or more images.
	We begin by introducing the IR and the MIR and motivate these features in an information theoretical context as the ratio of the self-information of an intensity level over the information contained over the pixels of the same intensity. 
	Notably, the relationship of the IR and MIR with  the image entropy and mutual information, classic  information measures, are discussed.
	Finally, the effectiveness of these features is tested through feature extraction over INRIA Copydays datasets and feature matching over the Oxford’s Affine Covariant Regions.
These numerical evaluations validate the relevance of the IR and MIR in practical computer vision tasks.
\end{abstract}

\begin{IEEEkeywords}
Computer vision; Entropy; Mutual Information; Feature counting; Feature Matching; 
\end{IEEEkeywords}
\section{Introduction}
\label{sec:Introduction}

Advances in computer vision and image processing methods  have relied on information theory as a powerful mathematical tool to determine the statistical variability of a set of images. 
Several saliency criteria have been proposed in the literature inspired by information theoretical concepts, such as the Kullback-Leibler divergence \cite{jagersand1995saliency} and the Shannon entropy \cite{kadir2001saliency,sponring1996entropy}.
In the following, we define two new image features motivated by information theoretical concepts: the Information Ratio (IR) and the Mutual Information Ratio (MIR).
The IR is obtained as the self-information of a intensity of image over the information content of the pixels of the same intensity.
The MIR is defined in  a similar manner but with respect to a pair of intensity levels across two images.
We show the effectiveness of these proposed image features through standard computer vision tasks: feature counting and feature matching.

\subsubsection*{Related results}
In computer vision, local image features such  as edges, corners and lines are  extracted from an image to empower that mid- and high-level vision tasks, such as image registration, motion tracking, and 3-D reconstruction \cite{ruiz2009information}. 
Generally speaking, image features are either handcrafted or obtained through deep learning.
 
Conventional handcrafted feature-extracting algorithms are  FAST \cite{rosten2006machine} and Oriented FAST and Rotated BRIEF (ORB) \cite{rublee2011orb}, Speeded Up Robust Features (SURF) \cite{bay2008speeded}, and KAZE (meaning ``wind'' in Japanese) \cite{alcantarilla2012kaze}.
Among deep-learning inspired feature-extracting algorithms, we mention FASTER \cite{rosten2008faster} and  V-FAST  \cite{yu2010real}.
All the algorithms  above produce specific descriptor and extract the different count of local features from the images.
To evaluate the efficiency of the feature-extraction algorithms above,  two criteria are usually considered: the count of extracted features and computational complexity \cite{bay2008speeded,alcantarilla2012kaze,fan2015local}.
%
Although the image features count is a helpful criterion for comparing two different algorithms, it singly and generally provides no guarantees on the quality of these features or the expected number of features extracted for different sets of images. Thus, it needs to estimate a reference value for the count before feature extracting.
%
Similarly, the computational complexity is a relative criterion in determining the performance of a feature extraction algorithm.
Accordingly, the selection and the configuration of a specific feature extraction algorithm  on a specific image  dataset is depended on the human decision and not a firm mathematical basis.
\begin{figure}[h]
	\includegraphics[width=\linewidth]{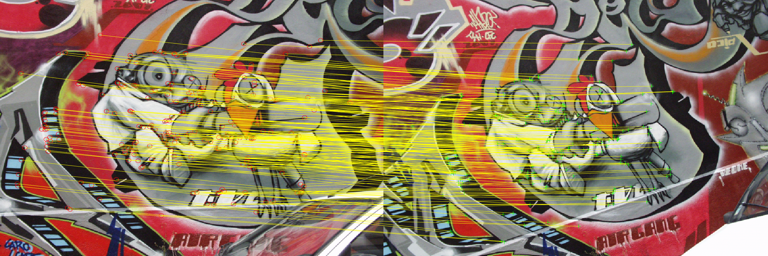}
	\caption{Local features matching by the ORB algorithm \cite{rublee2011orb} on the gray-scale image of a graffiti.}
	\label{fig:matching}
\end{figure}
Image features are often used for feature matching which is the procedure in  which features are matched in two consecutive images, usually in gray-scale image or in one of the image channels. 
Feature matching, also depicted in Fig. \ref{fig:matching},  is used in advanced computer vision tasks, such as depth estimation, 3D reconstruction, and motion capture.

For image matching,  that is the problem of determining object features in two different images of the same scene,  the curve of ``precision versus recall'' is sometimes used to assess local descriptors and features. 
In this curve, the ``precision'' argument corresponds to the ratio of correct over total matches, while the ``recall'' argument to the portion of 
features that are matched between the original image and its transformation. Another tool  to evaluate feature matching algorithms for a given dataset is the average precision, as described in \cite{fan2015local}.

\subsubsection*{Contributions}
In this paper, the Information Ratio (IR) and the Mutual Information Ratio (MIR) features are introduced as image features across a single and two (or more) images.
We argue for the usefulness of the proposed image features through three sets of numerical evaluations: (i) by computing the IR features in  two conventional image datasets: University of Oxford’s Affine Covariant Regions \cite{oxf} and INRIA Copydays \cite{inria}, (ii) by evaluating the feature distance for first dataset to evaluate the feature matching performance using MIR features, and (iii) a pre-processing method based on the IR feature evaluated on the second dataset. 

The lower bounds of the IR and the MIR as the functions of the entropy and the mutual information are also provided. 
This approximation is particularly useful to reduce the computational complexity in determining the features count.

From a high level perspective, our motivation in resorting to an information theoretical approach to the feature matching problem is in determining the fundamental answer to the following questions:  
(i) How many feature points exist in a given image, regardless of how features are described? and (ii)   How many common features can be determined among two given images?

To answer these questions, we resort to a statistical formulation of the feature extraction problem and the feature matching problem which, we hope, will find general applicability.
%
%
%
\subsubsection*{Notation}
In the remainder of the paper we use the following notation. 
We define $[n:m] \triangleq \{n,n+1, \ldots, m-1,m \} \subset \Nbb$. When $n=0$, we use the simpler notation $[m]$.
Random Variables (RVs) are indicated with capital letters, i.e. $X$.   
Random Vectors (RVs) are indicated with bold capital letters, i.e. $\Xv$.
%
The hat symbol is used to indicate the sample version and the symbol tilde for histogram versions of a random object.  
Logarithms are taken in any base.

\smallskip
The code for the numerical simulations in the paper can be found at the following web address: \url{https://github.com/AliKhajegiliM/IR-and-MIR}.

\section{Problem Formulation}
\label{sec:Problem Formulation}
\subsection{System Model}
\label{sec:System Model}
In the standard RGB (Red, Green, and Blue), an image is described through a matrix of dimension $N \times M$  where each element is referred to as a pixel.
Each pixel is itself a vector in $[2^D-1]^3$ where $D$ is the color depth of an image and each dimension represents one of the following colors: red, green, and blue which we indicate as $\Rv,\Gv$, and $\Bv$, respectively.
We assume that a set of $L$ images $\Xv=\{\Xv_1, \ldots, \Xv_{L}\}$ is drawn from the distribution $P_{\Xv}$  where $\Xv_t \in [N]\times [M] \times [2^D-1]^3$ (that is pixels 2-D position plus 3D color vector).   In the following, we refer to each $\Xv_t$ as a \emph{frame}. For a given $\Xv_t$,  $\Xv_t(\Cv)$ let $ \Cv \in \{  \Rv, \Gv , \Bv\}$ be the 2-D vector corresponding to the color $\Cv$; we refer to  $\Xv_t(\Cv)$ as the \emph{component image} of the color $\Cv$ or the \emph{ $\Cv$ channel} of the image $\Xv_t$. 
 Finally $X_t(\Cv,n,m)$ for $(n,m) \in [N]\times [M]$ represents the intensity of the pixel $(n,m)$ for the color $\Cv$ at frame $t$.
Generally speaking, as the $\Rv$, $\Gv$, and $\Bv$ component images are correlated, the elements of the random vector are also correlated. 
This also holds for the pixels in the component images. 
In the following, we will be concerned with estimating the image histogram of the frame $\Xv_t(\Cv)$ with the distribution $P_{\Xv(\Cv)}$: 
\ea{
	\tilde{h}_i(\Cv) \triangleq \sum_{n,m \in [N]\times [M]} 1_{ \{X_t(\Cv,n,m)=i \}},  \ \forall i \in [2^D-1]
	\label{eq:count Xo}
}
where $\ones_{\{C\}}$ is the indicator function of the condition $C$.

The sample self-information of the $i^{\rm th}$ intensity level is then estimated as  $-\log \lb 	\tilde{h}_i(\Cv)/NM\rb \ [\rm{nats}]$  and the  sample entropy as the 
\ea{
	\Hh(\Cv) \triangleq -\sum_{i \in [2^D-1]} p_i(\Cv) \log \lb p_i(\Cv) \rb  \ [\rm{nats}],
	\label{eq:sample entropy}
}
for $p_i(\Cv)=	\tilde{h}_i(\Cv)/(NM)$.
In the following we refer to $p_i(\Cv)$ as  the sample probability mass function of $\Xv_t(\Cv)$.
The definitions in \eqref{eq:count Xo} and \eqref{eq:sample entropy} can be extended to two sets of consecutive frames 
as follows
\ea{
	\Xv^2(\Cv) = [\Xv_1(\Cv) , \Xv_2(\Cv)],
	\label{eq:def Xo 2}
}
Accordingly, the joint histogram of $\Xv^2(\Cv)$ with the distribution $P_{\Xv^2(\Cv)}$ can be estimated analogously to \eqref{eq:count Xo} as follows
\ea{
	\tilde{h}_{i,j}(\Cv) \triangleq \sum_{n,m \in [N]\times [M]} 1_{ \{X_1(\Cv,n,m)=i, \ X_2(\Cv,n,m)=j \}}, 
	\label{eq:count Xo 2}
}
for $p_{i,j}(\Cv)=\tilde{h}_{i,j}(\Cv)/(NM)$ and $i,j \in [2^D-1]$.
The sample joint entropy of $\Xv_1(\Cv)$ and $\Xv_2(\Cv)$ is defined as 
\ea{
	\Hh(\Cv,\Xv_1,\Xv_2)=-\sum_{{i,j} \in [2^D-1]}p_{i,j}(\Cv)\log(p_{i,j}(\Cv)),
	\label{eq:entropy}
}
and the sample mutual information as 
\ea{
	\Ih(\Cv,\Xv_1;\Xv_2)=\Hh(\Cv,\Xv_1)+\Hh(\Cv,\Xv_2)-\Hh(\Cv,\Xv_1,\Xv_2).
	\label{eq:mutual information}
}
\begin{remark}
As defined by Shannon in \cite{shannon1948mathematical}, the entropy conceptually captures the information, or better \emph{variability}, in a random source.
Although there already various image processing algorithms have been inspired by information theoretical concepts, see \cite{ruiz2009information}, our approach is substantially different from the other approaches in the literature.
We introduce the IR and the MIR as novel information measures for images and sets of images, respectively: these measures are substantially different from previously-investigated information measures. 
\end{remark}
\begin{remark}
Due to space limitations, a number of extensions to our analysis are not pursued here. 
For instance, (i) we don't consider the effect of correlation among the channels, (ii) the MIR for more than two images, and (iii) extensions to other image color models, such as the  CMYK color model.
Such extensions are left for future research. 
\end{remark}
\section{The Information Ratio (IR) and the Mutual Information Ratio (MIR) Features}
\label{sec:The Information Ratio (IR) and the Mutual Information Ratio (MIR) Features}

In this section we define two new image features based on the quantities introduced in Sec. \ref{sec:System Model}. 
Let us begin with the IR feature. 
Note that we have $\tilde{h}_{i}(\Cv)$ pixels with self-information $-\log(p_{i}(\Cv))$, so that we have a cumulative information $-\tilde{h}_i(\Cv) \log p_i(\Cv) \ [\rm {pixel-nats}]$. On the other hand, the chance of selecting each of these $\tilde{h}_i(\Cv)$ points under a uniform distribution is ${1}/{\tilde{h}_i(\Cv)}$ with a self-information $\log(\tilde{h}_i(\Cv)) \ [\rm{nats}]$. 
Using the definition of the random vectors, the ratio  between the cumulative information and the self-information under uniform distribution is
\ea{
R_i(\Cv) & =  \f {-\log \lb P_{\Xv(\Cv)}(i)\rb }{ \log \lb NM P_{\Xv(\Cv)}(i) \rb}.
\label{eq:r_i}
}
We refer to \eqref{eq:r_i} as information ratio of the intensity level $i \in [2^D-1]$. 
Next, we define the expected value of \eqref{eq:r_i}  as the {average information ratio} over the channel pixels, that is 
\ea{
R(\Cv,X) = NM\times\Ebb_{\Xv(\Cv)} \lsb R_i(\Cv) \rsb \ \ [ \rm {pixel}].
\label{eq:average information ratio}
}
In the following, we use the histogram version of the quantity in \eqref{eq:average information ratio} as an image feature.
let ample version of \eqref{eq:r_i} be
\ea{
r_{i}(\Cv,\Xv) & = 
\lcb \p{
 \f{- \log \lb p_{i}(\Cv)\rb }{\log \lb \tilde{h}_{i}(\Cv)\rb} & \tilde{h}_{i}(\Cv)>1 \\
 0 & \rm{ otherwise} 
}
\  \ 
\rnone
}
so that the histogram version of  \eqref{eq:average information ratio} is
\ea{
r(\Cv,\Xv)  & =  \sum_{i \in [2^D-1]} \tilde{h}_{i}(\Cv)  r_{i}(\Cv,\Xv) \  \ [ \rm {pixel}].
\label{eq:information ratio}
}
We term the quantity in \eqref{eq:information ratio} as the \emph{ Information Ratio (IR)} image feature.\\
The MIR feature is defined in an analogous way as the IR image feature in \eqref{eq:information ratio} but across two (or more images). 
First, we define the self-information ratio of two intensity level $i$ an $j$ in the frames $\Xv_1(\Cv)$ and $\Xv_2(\Cv)$ as
\ea{
	M_{i,j}(\Cv,\Xv_1;\Xv_2) & =  \f {\log \lb \dfrac{P_{\Xv^2(\Cv)}(i,j)}{P_{\Xv_1(\Cv)}(i)P_{\Xv_2(\Cv)}(j)} \rb }{\log \lb NM P_{\Xv^2(\Cv)}(i,j) \rb}.
	\label{eq:cmutu}
}
for the images  $\Xv_1$ and $\Xv_2$ as in \eqref{eq:def Xo 2}.
The definition in \eqref{eq:cmutu} is intuitively  motivated as follows.
Consider  two images $\Xv_1$ and $\Xv_2$,
the self  mutual information of each pair $(i,j)$ of intensities  in two image channels is $\log(p_{i,j}(\Cv))-\log(p_i(\Cv))-\log(p_j(\Cv))$. 
Also, $\tilde{h}_{i,j}(\Cv)$ pixels with intensity $i$ in the first image are  ``matched'' with pixels of intensity $j$ in the second image.
%
%
What is the probability of the random selection of involved pixels in the intensity pair (i,j) as a matched pixels? This probability of this event is  ${1}/ \tilde{h}_{i,j}(\Cv)$. 
This means that the self-information in this random selection is $\log(\tilde{h}_{i,j}(\Cv))$. 
Accordingly, the ratio  in \eqref{eq:r_i} for the pair of intensities $(i,j)$ in two images is as in \eqref{eq:cmutu}.
The  \emph{average mutual information ratio} between two images $\Xv_1$ and $\Xv_2$ over all pixels is defined as
	\ea{
		M (\Cv,\Xv_1;\Xv_2) & = NM \times \Ebb_{\Xv^2} \lsb M_{i,j}(\Cv,\Xv_1;\Xv_2) \rsb 
		\label{eq: muratio}
	}
The  histogram version of the average mutual information ratio  in  \eqref{mutaulratio} is defined analogously to  \eqref{eq:information ratio} as

\ea{
m( \Cv,\Xv_1;\Xv_2) & =\sum_{ \{j,i\} \in [2^D-1]} \tilde{h}_{i,j}(\Cv)\dfrac{\log \lb \dfrac{p_{i,j}(\Cv)}{p_{i}(\Cv)p_{j}(\Cv)}\rb}{\log(\tilde{h}_{i,j}(\Cv))}.
\label{mutaulratio}
}
We define the quantity in \eqref{mutaulratio} as the \emph{Mutual Information Ratio (MIR)} image feature. 
From a high-level perspective, the IR in \eqref{eq:information ratio} captures the information contained in a singular pixel of intensity $i$ as compared to all other pixels with the same intensity. 
A further interpretation of the IR is as follows: the image histogram is not a sufficient statistic of the original image, as the position information is not preserved. 
The uncertainty on the intensity of a pixel in the image is related to  the abundance of pixel of that intensity $i$.
Similarly, the MIR in \eqref{mutaulratio} provides a estimate of the information contained in a pair of the pixels of intensity $(i,j)$ positioned in the two images. 
The next theorems yield bounds on the IR and MIR  for a given image channel $\Cv$ on one frame and two.
\subsection{Some useful inequalities} 
The next inequalities are presented without proof, due to space constraints.
These inequalities are meant to provide useful bounds between the IR and the entropy, and between the MIR and the mutual information. 
\begin{thm}
			 \label{th:IR geq H}
	A lower bound on $r(\Cv,\Xv)$ is obtained as a function of $\Hh(\Cv)$ as
	\ea{
		 r(\Cv,\Xv) \geq \dfrac{NM}{\log(NM)}\Hh(\Cv) \geq 0,
		 \label{eq:IR geq H}
		\\ \nonumber \text{This lower bound is termed as LIR.}
		}
\end{thm}
The result in Th. \ref{th:IR geq H} shows that the image entropy, as defined in \eqref{eq:sample entropy}, is  a lower bound to the IR image feature. 
Note that the quantity in \eqref{eq:sample entropy} is consummately referred to as image entropy and is a common measure of variability of an image in computer vision.

\begin{thm}
	\label{th:MIR geq MI}
	A lower bound on $m(\Cv,\Xv_1;\Xv_2)$ is obtained as a function of $\Ih(\Cv,\Xv_1;\Xv_2)$ as 
	\ea{
		m(\Cv,\Xv_1;\Xv_2) \geq \dfrac{NM}{\log(NM)}\Ih(\Cv,\Xv_1;\Xv_2)\geq 0,
			\\ \nonumber \text{This lower bound is termed as LMIR.}
	}
\end{thm}
Note that both the IR and the MIR are positive defined. 
The sample mutual information in \eqref{eq:mutual information}, as the sample entropy in \eqref{eq:entropy}, is a commonly-used measure in computer vision to quantify the similarity among two images. 

{
	\begin{lem}
		\label{lem:up bound}
		When $D \leq \dfrac{\log(NM)}{2\log(2)}$, then 
		\ea{
			r(\Cv,\Xv)\ \text{and} \ m(\Cv,\Xv_1;\Xv_2) \leq NM
		}
	\end{lem} 
	Generally, customary cameras, the resolution of daily captured images, and color mapping technologies satisfy the condition in Lem. \ref{lem:up bound}.  
}
%
%

\newtheorem{corollary}{Corollary}[section]
\begin{corollary}
\label{cor:simplerelation}
	The following inequalities hold
	\ea{
		-\sum_{ \{j,i\} \in [2^D-1]}\tilde{h}_{i,j}(\Cv)\dfrac{\log(p_{i}(\Cv))}{\log(\tilde{h}_{i,j}(\Cv))}
		\geq r(\Cv,\Xv_t) \geq 0, 
	\label{simplerelation}
}
for $t\in \{1,2\}$ and	where $r(\Cv,\Xv_t)$ is the IR of the channels $\Cv$ for  the first and second images.
\end{corollary}
Cor. \ref{cor:simplerelation} shows how the IR feature is upper bounded by a function of the joint histogram of the two images.
%

The results in Th. \ref{th:IR geq H},  Th. \ref{th:MIR geq MI}, Lem. \ref{lem:up bound} and Cor. \ref{cor:simplerelation} are presented here to highlight some useful properties of the IR and the MIR image feature. 
In particular, these results highlight the IR and the MIR can be approximated. 
\section{Numerical Experiments}
\label{sec:Numerical Experiments}
In this section we present three sets of numerical evaluations to validate the use of IR and the MIR as image features. 
The IR feature is used for feature points count in a single image, while the MIR is used for feature matching across two images. 
\subsubsection*{Experiment Setup}
For the numerical evaluations, a  computer system  with Processor Intel(R) Core(TM) i5-6200U CPU @ 2.30GHz, 2.40GHz, 8.00 GB RAM is used. 
MATLAB software is  used for the experiments, in particular the MATLAB computer vision toolbox implementations of the ORB, SURF(64D), and KAZE is used for extracting the image feature points, and \textit{matchFeatures} function used for matching them.
\subsubsection*{Datasets}
\begin{figure*}[h]
	\includegraphics[width=1\linewidth,center]{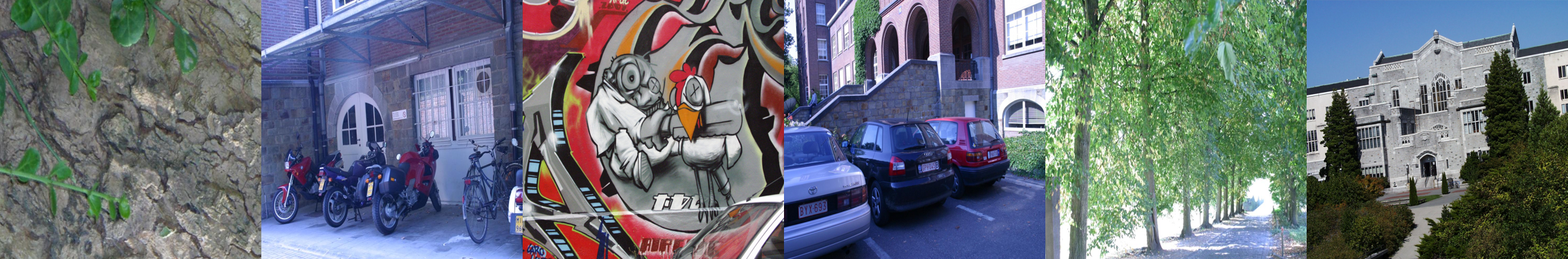}
	
	\caption{\textbf{Oxford’s Affine Covariant Regions}: \textit{from left to right}, bark ($512\times765$) for rotation, bikes ($700\times1000$) for blur scene, graffiti ($640\times800$) for orientation and field of view variation, Leuven ($600 \times 900$) for intensity variation, trees ($700\times1000$) for blur scene, and UBC ($640\times800$) for JPEG compression.
	}
	\label{fig:oxf}
\end{figure*}

We evaluate the IR and the MIR image feature performance over  two conventional image datasets: the University of Oxford’s Affine Covariant Regions \cite{oxf} and INRIA copydays \cite{inria}.
In the first dataset, there exist six type of images with six images for each type: for evaluating specific purposes, Fig. \ref{fig:oxf} provides some illustrative examples. 
The second dataset is comprised of  157 random images from different scenes, places, and creatures.

\subsubsection*{Overview}
ORB, SURF, and KAZE algorithms are applied to the datasets to obtain the per-frame(channel) features and matched features of two consecutive frames(channels), respectively.
Simulations are performed over all three channels of an image, i.e. $\Cv \in \{\Rv, \Gv ,\Bv\}$.
For brevity, only section the results of Red channel are reported here.
The performance in the remaining two channels substantially confirm our conclusions. 
Evaluations are repeated for multiple images in the same category in the dataset.  
\subsubsection*{Image Feature distance}
{
The KAZE, SURF, and ORB extract general local image features. 
	Generally speaking, a local image feature is a neighborhood, containing some pixels of the same intensity level,  which identifies salient points such as edges, lines, corners, textures, and so forth.
	The images in the datasets have  256-level pixels: to avoid identifying the same image feature multiple times, we consider two features to be valid when they have minimum distance greater than a chosen threshold $d$.
	This is equivalent of reducing the histogram levels by merging contiguous intensity levels. 
	For instance, sub-sampling the intensity levels to 128 values coincides to a minimum feature distance of $d=2$.
	Note that, in human perception, the people's eyes are sensitive to the  different feature distances, according to the individual visual psychology \cite{elliot2014color}.}
In the remainder of the section, results are listed for five values of the feature distance: $d \in \{1,8\}$.
\subsubsection*{IR feature evaluation}

\begin{figure*}[h]
	\hfill
	\begin{subfigure}[Title A]{0.49\textwidth}
		\includegraphics[width=0.95\textwidth,center]{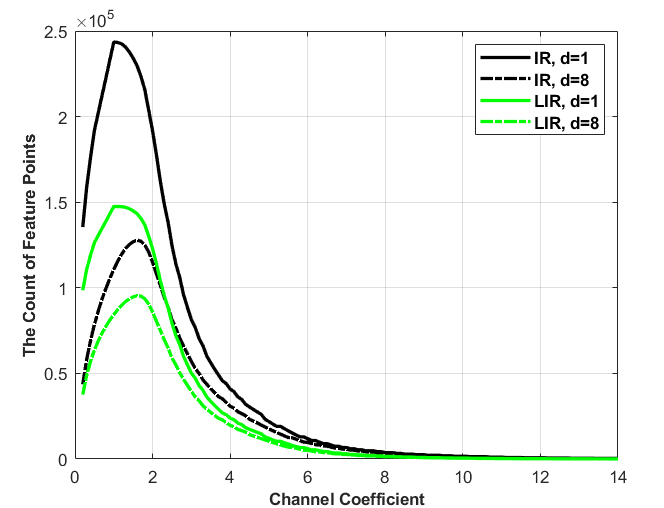}
		\caption{The IR feature count and its lower bound(LIR) as a function of the brightness level $K$. This shows the dependency between feature distance and the peak of feature counts curve.}
		\label{fig: mir curve}
		\hfill
	\end{subfigure}
	\begin{subfigure}[Title B]{0.49\textwidth}
		\includegraphics[width=0.9\textwidth,center]{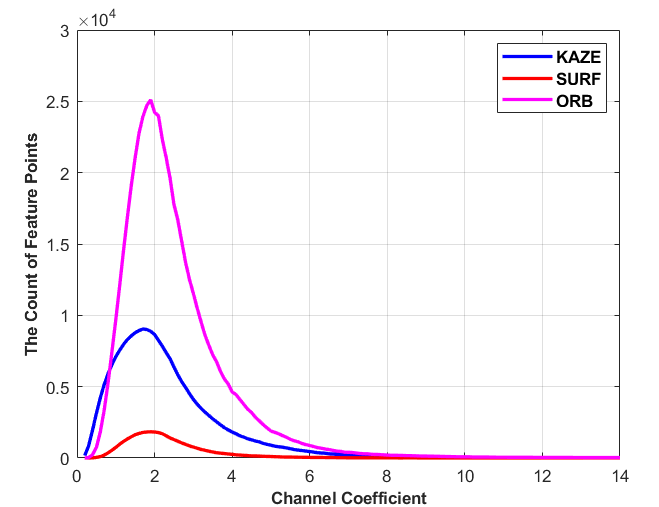}
		\caption{The KAZE, SURF, and ORB feature count as a function of the brightness level $K$.}
		\label{fig: surf kaze orb}
		\hfill
	\end{subfigure}
	\caption{The IR (left), KAZE, SURF, and ORB (right) features count for varying levels of brightness $K$ as obtained from the bark data in the University of Oxford's Affine Covariant Regions.}
	\label{fig: curves}
\end{figure*}

We evaluate the IR image measure by obtaining the IR feature count with varying levels of image brightness. 
To do so, we consider a varying brightness level $K$, multiply the intensity over each channel by the intensity level $K$ and obtain the IR feature count for the chosen level of $K$.
This experiment is performed on both datasets: a part of the results is presented in Table \ref{tab: feat co}.
The results are also presented in  Fig. \ref{fig: curves} in a visual manner.
\begin{table}[h]
	\centering
	\begin{tabular}{|c|c|c|c|c|c|c|}
		\hline
		\multirow{3}{*}{Dataset} & \multicolumn{6}{c|}{Mean values (Precision : *E03)}                          \\ \cline{2-7} 
		& \multicolumn{2}{c|}{IR} & \multicolumn{2}{c|}{LIR} & \multirow{2}{*}{KAZE} & \multirow{2}{*}{ORB} \\ \cline{2-5}
		& d=1        & d=8        & d=1         & d=8        &                       &                      \\ \hline
		Graffiti                 & 348.3      & 168.4      & 205.9       & 125.8      & 8.6                   & 12.7                 \\ \hline
		Bark                     & 248.5      & 113.8      & 149.5       & 86.3       & 7.4                   & 12.5                 \\ \hline
		Bikes                    & 447.7      & 215.1      & 270.1       & 162.2      & 5.3                   & 4.5                  \\ \hline
		Leuven                   & 306.9      & 137.0      & 187.9       & 103.4      & 4.5                   & 7.0                  \\ \hline
		Trees                    & 442.6      & 214.2      & 267.3       & 161.7      & 15.2                  & 46.3                 \\ \hline
		UBC                      & 302.6      & 158.5      & 179.4       & 117.9      & 8.6                   & 23.9                 \\ \hline
	\end{tabular}
\caption{A comparison of IR feature and KAZE and ORB algorithms for feature counting for various feature distance $d$ over the red channel. 
	The lower bound in Th. \ref{th:IR geq H} (indicated as LIR) is also reported.
}
\label{tab: feat co}
\end{table}
\subsubsection*{MIR feature evaluation}
For the MIR feature, we compare the matched points in two successive frames to within a desired feature distance $d$ as compared to KAZE, SURF, and ORB.
To verify the predictive value of the lower bound in Th. \ref{th:MIR geq MI}, we also evaluate this lower bound, a part of results is reported in Table \ref{tab: match co}. 

\begin{table}[h]
	\centering
	\begin{tabular}{|c|c|c|c|c|c|c|}
		\hline
		\multirow{3}{*}{Dataset} & \multicolumn{6}{c|}{Mean values (Precision : *E03)}                                \\ \cline{2-7} 
		& \multicolumn{2}{c|}{MIR} & \multicolumn{2}{c|}{LMIR} & \multirow{2}{*}{KAZE} & \multirow{2}{*}{ORB} \\ \cline{2-5}
		& d=1         & d=8        & d=1         & d=8         &                       &                      \\ \hline
		Graffiti                 & 10.5        & 2.8        & 9.1         & 2.0         & 0.01                  & 0.3                  \\ \hline
		Bark                     & 14.2        & 2.1        & 8.3         & 1.3         & 0.4                   & 1.4                  \\ \hline
		Bikes                    & 136.4       & 69.8       & 61.5        & 49.5        & 3.1                   & 1.3                  \\ \hline
		Leuven                   & 111.6       & 43.6       & 48.1        & 30.4        & 2.6                   & 1.8                  \\ \hline
		Trees                    & 56.9        & 16.8       & 21.4        & 11.9        & 2.8                   & 0.5                  \\ \hline
		UBC                      & 187.7       & 84.6       & 69.8        & 56.5        & 3.8                   & 4.8                  \\ \hline
	\end{tabular}

\caption{A comparison of MIR feature and KAZE and ORB algorithms in the feature matching task for various feature distance $d$ over the red channel. 
	The lower bound in Th. \ref{th:MIR geq MI} (indicated as LMIR) is also reported. }
\label{tab: match co}
\end{table}
\subsubsection*{IR based optimization method}
This experiment which is devised to analyze the IR applicability, shows that the maximum count of the extracted features is not necessarily for $K=1$, it is showed in \figurename\ref{fig: surf kaze orb}. This means that there is a coefficient $K$ which maximizes the extracted features count of the applied image. We aim to find this optimizer coefficient by virtue of the IR measure. The optimizing algorithm procedure is reported in Alg.\ref{alg} and the effectiveness of using the optimized images for feature extracting over the INRIA Copydays dataset is shown in Table \ref{tab: optimization}. The computational complexity of finding $K_{Optimizer}$ on average is $0.0015^{sec}$, which is derived from multiple tests on the dataset.

	\begin{algorithm}[]
		\centering
		\caption{The IR-based Optimizer}	\label{alg}
		\begin{algorithmic}[1]
			\Procedure{:}{}
			\State $d=8 \ \gets \text{Set the desired feature distance}$.
			\State $m \gets$ Compute the input channel($\Xv(\Cv)$) mean
			\State \textbf{For}\  $K \in [0.9, \frac{255}{m}],\ step=0.1$:
			\State \ \ $h_{i}^{K, d}(\Cv)\gets$ Compute  the histogram of $K\times \Xv(\Cv)$
			\State \ \ $IR^{K,d} \gets$ Compute the IR value of $h_{i}^{K,d}(\Cv)$
			\State \ \ \textbf{If}\  $IR^{K, d}-IR^{K-0.1,d} \leq 0$:
			\State \ \ \ $K_{Optimizer} \gets K-0.1$, \textbf{Break}
			\State \ \ \textbf{End If}
			\State  \textbf{End For}
			\State Use  $K_{Optimizer} \times \Xv(\Cv)$ instead of $ \Xv(\Cv)$
			\EndProcedure
		\end{algorithmic}
	\end{algorithm}
\begin{table}[]
	\centering
	\begin{tabular}{|c|c|c|c|}
		\hline
		\multirow{2}{*}{Algorithm} & \multicolumn{2}{c|}{Extracted Features Count} & \multirow{2}{*}{\begin{tabular}[c]{@{}c@{}}Improvement\\ Rate\end{tabular}} \\ \cline{2-3}
		& $R$                         & $K_{Optimizer}\times R$                               &                                                                             \\ \hline
		KAZE                       & 12.34                     & \textbf{13.94}                    & \textbf{12.99\%}                                                              \\ \hline
		ORB                        & 33.70                      & \textbf{42.43}                     & \textbf{25.91\%}                                                             \\ \hline
		SURF                       & 2.13                      & \textbf{2.78}                     & \textbf{30.93\%}                                                             \\ \hline
	\end{tabular}
\caption{The computational comparison of feature extracting performance on the optimized image and the original image. The applied algorithms are the KAZE, SURF, and ORB over the red channel(the reported counts precision: *E03). }
\label{tab: optimization}
\vspace{-0.5 cm}
\end{table}

\section{Discussion}
\label{sec:Discussion}
The image  channel histogram,  on a first approximation, is invariant to orientation and rotation, it is also somewhat robust to variation in intensity, quality and field of view. 
For this reason, the  IR and the MIR image features also possess these characteristics, thus making these image features particularly attractive for practical applications. 
In the concept of scale variation, it is worth noting that the count of feature points is depended on the scale of image. In this study, the devised manners are used for consecutive frames with the same size, but for frames with different sizes a helpful way is up-sampling or down-sampling method to make them the same in size. 
The IR and the MIR image features, are fundamentally linked to the image entropy and image mutual information which are fundamental measures of variability and dependence among images, respectively.
It is worth nothing that in image processing, the entropy and the mutual information are global, rather than local features.
Accordingly, the lower bounds in Th. \ref{th:IR geq H} and Th. \ref{th:MIR geq MI} are rather useful lower bounds to the the IR and the MIR performance.
As such, these bounds can be used to predict this performance before calculation, which can be rather intensive. 
In \figurename\ref{fig: curves}, it is shown that feature extraction algorithms follow the same curve as the IR feature. 
This, intuitively, expresses that with varying $K$, the features in an image change and the IR appropriately captures this variation. The curves related to $d=8$ reveal this fact that the best coefficient $K$ is not \textit{one} as the algorithms show in \figurename\ref{fig: surf kaze orb}. 
Based on the reported results in Table \ref{tab: feat co}, it is clear that the other feature matching in the literature, i.e. KAZE, SURF, and ORB,  can extract 
an amount of features lower than the IR and LIR features. 
Since the goal in many computer vision tasks uses features as a starting point, having a larger amount of starting feature points can potentially yield a better result.
For this reason, we conclude that the large number of the IR feature can potentially enable better performance, although this claim requires further investigation. 
Feature matching is indeed among these higher-level computer vision tasks. 
Let us next discuss the result as reported in Table \ref{tab: match co}.
Only the ORB algorithm and for $d\geq8$,  matches more  features than the LMIR feature and is close to the IR.
In almost other distances the MIR yield more matches that the KAZE and ORB.
Also note how closely the LMIR predicted the matches obtained through the MIR features. 
By comparing the Table \ref{tab: match co} and Table \ref{tab: feat co}, we note that the count of matching MIR feature is much lower than the number of total IR features.
%
The effectiveness of the IR feature is shown computationally in Table \ref{tab: optimization}, the count of extracted features through the KAZE, SURF , and ORB algorithms are maximized by finding the corresponding $K_{Optimizer}$ based on the IR feature in a flash of second.

In conclusion, we argue that the results presented here
show that the IR and the MIR features have the potential improving the preformance of a number of computer vision. 
Future research direction will focus in determining the computational complexity of relevant algorithms operating on larger dataset. 
Although present algorithms, such as the KAZE, SURF, and ORB, cannot produce as many features as the IR and the MIR features, further research is necessary to better determine the potential of these novel features.
\section{Conclusion}
\label{sec:Conclusion}
Image feature extraction and matching is a prerequisite to high-level computer vision tasks. 
For this reason, obtaining a large set of features and matches leads to more effective algorithms.
In this paper, two new image features  are proposed: the Information Ratio (IR) and the Mutual Information Ratio (MIR) features. 
The IR feature is used to produce a large count of local image feature points.
The MIR feature  is used to match local feature  across two (or more) consecutive frames. 
We show that  the sample entropy and the sample mutual information are lower bound on IR and MIR, respectively. 
Therefore, the optimization methods which need more image information, i.e. exactly image feature points, can use IR and MIR instead of the entropy and mutual information as objective functions. One method to maximize the detectable features based on the IR feature is devised which it outperforms significantly applied feature extraction algorithms.
We also compare the IR and the MIR features with other image features proposed in the literature: the ORB, KAZE, and SURF features. 
More precisely, we compare the IR with these features in two tasks: total features per image and feature matching across frames in two datasets. 
These numerical evaluations are only a partial characterization of the potential of these two measures. 
Further research is necessary to more thoroughly characterize the task better suited for the IR and the MIR image features. 

\bibliographystyle{IEEEtran}
\bibliography{IWCIT_bib}

\begin{thebibliography}{10}
\providecommand{\url}[1]{#1}
\csname url@samestyle\endcsname
\providecommand{\newblock}{\relax}
\providecommand{\bibinfo}[2]{#2}
\providecommand{\BIBentrySTDinterwordspacing}{\spaceskip=0pt\relax}
\providecommand{\BIBentryALTinterwordstretchfactor}{4}
\providecommand{\BIBentryALTinterwordspacing}{\spaceskip=\fontdimen2\font plus
\BIBentryALTinterwordstretchfactor\fontdimen3\font minus
  \fontdimen4\font\relax}
\providecommand{\BIBforeignlanguage}[2]{{%
\expandafter\ifx\csname l@#1\endcsname\relax
\typeout{** WARNING: IEEEtran.bst: No hyphenation pattern has been}%
\typeout{** loaded for the language `#1'. Using the pattern for}%
\typeout{** the default language instead.}%
\else
\language=\csname l@#1\endcsname
\fi
#2}}
\providecommand{\BIBdecl}{\relax}
\BIBdecl

\bibitem{jagersand1995saliency}
M.~Jagersand, ``Saliency maps and attention selection in scale and spatial
  coordinates: An information theoretic approach,'' in \emph{Proceedings of
  IEEE International Conference on Computer Vision}.\hskip 1em plus 0.5em minus
  0.4em\relax IEEE, 1995, pp. 195--202.

\bibitem{kadir2001saliency}
T.~Kadir and M.~Brady, ``Saliency, scale and image description,''
  \emph{International Journal of Computer Vision}, vol.~45, no.~2, pp. 83--105,
  2001.

\bibitem{sponring1996entropy}
J.~Sponring, ``The entropy of scale-space,'' in \emph{Proceedings of 13th
  International Conference on Pattern Recognition}, vol.~1.\hskip 1em plus
  0.5em minus 0.4em\relax IEEE, 1996, pp. 900--904.

\bibitem{ruiz2009information}
F.~E. Ruiz, P.~S. P{\'e}rez, and B.~I. Bonev, \emph{Information theory in
  computer vision and pattern recognition}.\hskip 1em plus 0.5em minus
  0.4em\relax Springer Science \& Business Media, 2009.

\bibitem{rosten2006machine}
E.~Rosten and T.~Drummond, ``Machine learning for high-speed corner
  detection,'' in \emph{European conference on computer vision}.\hskip 1em plus
  0.5em minus 0.4em\relax Springer, 2006, pp. 430--443.

\bibitem{rublee2011orb}
E.~Rublee, V.~Rabaud, K.~Konolige, and G.~Bradski, ``Orb: An efficient
  alternative to sift or surf,'' in \emph{2011 International conference on
  computer vision}.\hskip 1em plus 0.5em minus 0.4em\relax Ieee, 2011, pp.
  2564--2571.

\bibitem{bay2008speeded}
H.~Bay, A.~Ess, T.~Tuytelaars, and L.~Van~Gool, ``Speeded-up robust features
  (surf),'' \emph{Computer vision and image understanding}, vol. 110, no.~3,
  pp. 346--359, 2008.

\bibitem{alcantarilla2012kaze}
P.~F. Alcantarilla, A.~Bartoli, and A.~J. Davison, ``Kaze features,'' in
  \emph{European Conference on Computer Vision}.\hskip 1em plus 0.5em minus
  0.4em\relax Springer, 2012, pp. 214--227.

\bibitem{rosten2008faster}
E.~Rosten, R.~Porter, and T.~Drummond, ``Faster and better: A machine learning
  approach to corner detection,'' \emph{IEEE transactions on pattern analysis
  and machine intelligence}, vol.~32, no.~1, pp. 105--119, 2008.

\bibitem{yu2010real}
T.-H. Yu, T.-K. Kim, and R.~Cipolla, ``Real-time action recognition by
  spatiotemporal semantic and structural forests.'' in \emph{BMVC}, vol.~2,
  no.~5, 2010, p.~6.

\bibitem{fan2015local}
B.~Fan, Z.~Wang, F.~Wu \emph{et~al.}, \emph{Local image descriptor: modern
  approaches}.\hskip 1em plus 0.5em minus 0.4em\relax Springer, 2015, vol. 108.

\bibitem{oxf}
{Visual Geometry Group. (2004)}, ``{Affine Covariant Regions Datasets [Online],
  Available: http://www.robots.ox.ac.uk/{~}vgg/data}.''

\bibitem{inria}
{INRIA}, ``Copydays dataset [online], {Available:
  http://lear.inrialpes.fr/people/jegou/data.php},'' 2008.

\bibitem{shannon1948mathematical}
C.~E. Shannon, ``A mathematical theory of communication,'' \emph{Bell system
  technical journal}, vol.~27, no.~3, pp. 379--423, 1948.

\bibitem{elliot2014color}
A.~J. Elliot and M.~A. Maier, ``Color psychology: Effects of perceiving color
  on psychological functioning in humans,'' \emph{Annual review of psychology},
  vol.~65, pp. 95--120, 2014.

\end{thebibliography}

\appendix
\subsection{\textbf{The Proof of Theorem III.1}}
\begin{proof}The inequality $\log(\tilde{h}_{i}(\Cv)) \leq \log(MN), \forall i \in [2^D-1] $ holds based on the histogram definition.  Consider $\rt(\Cv,\Xv)$ as defined in \eqref{eq:information ratio}
	\eas{
		r(\Cv,\Xv)
		&=-\sum\limits_{i  \in [2^D-1]}p_{i}(\Cv)\dfrac{NM\log(p_{i}(\Cv))}{\log(\tilde{h}_{i}(\Cv))}\\
		&\geq -\sum_{i  \in [2^D-1]} p_{i}(\Cv)\dfrac{NM\log(p_{i}(\Cv))}{\log(NM)}\\
		&=\frac{NM}{\log(NM)}\Hh(\Cv),
	}
	which shows the desired result.
\end{proof}

\subsection{\textbf{The Proof of Theorem III.2}}
\begin{proof}Like above, $\log(\tilde{h}_{i,j}(\Cv)) \leq \log(MN), \forall i, j \in [2^D-1]$ holds. Consider $m(\Cv,\Xv_1;\Xv_2)$ as defined in \eqref{eq: muratio}
	\eas{
		m(\Cv,\Xv_1;\Xv_2)
		&=\sum_{ \{j,i\} \in [2^D-1]}
		\tilde{h}_{i,j}(\Cv)\dfrac{\log(\dfrac{p_{i,j}(\Cv)}{p_{i}(\Cv)p_{j}(\Cv)})}{\log(\tilde{h}_{i,j}(\Cv))}\\
		& \geq \sum_{ \{j,i\} \in [2^D-1]}	\tilde{h}_{i,j}(\Cv)\dfrac{\log(\dfrac{p_{i,j}(\Cv)}{p_{i}(\Cv)p_{j}(\Cv)})}{\log(NM)}\\
		&=\dfrac{NM}{\log(NM)}\Ih(\Cv,\Xv_1;\Xv_2),
	}
	which show the desired result.	
\end{proof}
\subsection{\textbf{The Proof of Lemma III.1}}
\begin{figure}[h]
	\includegraphics[width=\linewidth]{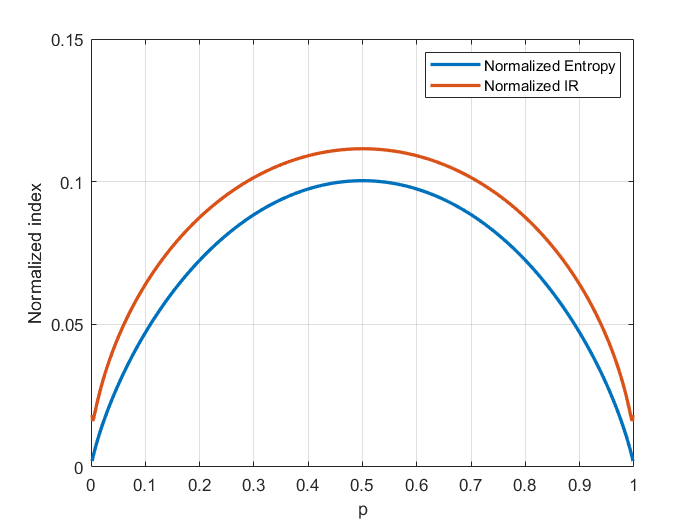}
	\caption{The curves of normalized IR and normalized entropy over a 2-symbol signal with the size of $NM=1000$. The probability of one symbol is considered $p$ which ranges within ($\frac{1}{NM},1-\frac{1}{NM}$).}
	\label{fig: app comp}
\end{figure}

\begin{proof}
	This lemma is proved in three steps. First, consider the Lagrange multiplier method by which we show that $r(\Cv,\Xv)$ maximizes through the uniform distribution. Thus, let $\lambda\in\Rv$ so that we aim to solve following equation(in this proof $p_i(\Cv)$ is replaced by $p_i$)
	\ea{
		\dfrac{ \partial }{\partial p_i}(r(\Cv,\Xv)+\lambda (\sum_{i  \in [2^D-1]}p_i \ - 1))=0,\  \forall i \in [2^D-1]
		\label{lem1}
	}
	by mathematical simplification,
	\ea{
		\frac{\lambda}{NM}=\dfrac{\log(p_i)\log(NMp_i)+\log(NM)}{(\log(NM)+\log(p_i))^{2}}	,\  \forall i \in [2^D-1]
		\label{lem2}
	}
	In fact, the equation \eqref{lem2} is a polynomial with the degree of two than $\log(p_i)$. The roots of this equation are depended on $\lambda$. Through solving that equation, one set of the desired responses for all $p_i$s is a constant amount, whether the equation has one response or two. 
	Furthermore, the constraint forces to make this amount equal for all $p_i$s. This means that the maximum of $r(\Cv,\Xv)$ occurs when the image is uniformly distributed.\\
	Second, consider a given image with the uniform distribution in order to find the maximum value of the IR.
	\eas{
		&\forall  i \in	[2^D-1], \ p_i=\frac{1}{2^D}\\
		&\Rightarrow r(\Cv,\Xv_{uniform})=\frac{NM  D \log2}{\log(NM)-D\log2}
	}
	So,
	\ea{
		D \leq \dfrac{\log(NM)}{2\log(2)}\Leftrightarrow  r(\Cv,\Xv_{uniform}) \leq NM
	} 
	It is worth noting that with respect to the first step, for all images $r(\Cv,\Xv) \leq r(\Cv,\Xv_{uniform})$. Accordingly, one part of the lemma is proved.\\
	Third, consider $M_{i,j}(\Cv,\Xv_1;\Xv_2)$ in \eqref{eq: muratio}, where is the maximum value of it?
	\eas{
		&\frac{\partial}{\partial p_{i,j}}M_{i,j}(\Cv,\Xv_1;\Xv_2)=0, \ \forall i,j \in [2^D-1]\\
		\Leftrightarrow &(\frac{1}{p_{i,j}}-\frac{1}{p_i}-\frac{1}{p_j})\log(p_{i,j})-\frac{1}{p_{i,j}}\log(\frac{p_{i,j}}{p_ip_j})=0\label{lem3} 
	}
	If two applied images are the same, $\Xv_1=\Xv_2$, the acceptable response of the equation \eqref{lem3} is obtained which leads to
	\ea{
		p_{i,j}=\lcb \p{ p_i=p_j  &	i=j
			\\
			0 & {otherwise}
		}\rnone
		\label{lem4}
	}
	Therefore, two equal consecutive frames cause to the maximum ratio. The following expressions show the upper bound on $m(\Cv,\Xv_1;\Xv_1)$. Put the probability mass function from \eqref{lem4}
	\eas{
		&m(\Cv,\Xv_1;\Xv_1)=\sum_{ \{j,i\} \in [2^D-1]}h_{i,j}(\Cv)M_{i,j}(\Cv,\Xv_1;\Xv_1)\\
		&=\sum_{i  \in [2^D-1]}\tilde{h}_i(\Cv)\frac{\log(p_i(\Cv))}{\log(\tilde{h}_i(\Cv))}
		=r(\Cv,\Xv_1)
	}
	Acordingly, similar to the first step $m(\Cv,\Xv_1;\Xv_1)$ maximizes when is uniformly distributed. Also, by analogy with the second step, 
	\ea{
		D \leq \dfrac{\log(NM)}{2\log(2)}
		\Leftrightarrow  m(\Cv,\Xv_{1,uniform};\Xv_{1,uniform}) \leq NM
	} 
	which results to
	\ea{
		m(\Cv,\Xv_{1};\Xv_{2}) \leq NM
	}
	The desired results yield in three steps. In order to make a visual sense the \figurename\ref{fig: app comp} is reported which shows the resemblance between the normalized IR feature(divided by $NM$) and  the normalized entropy (divided by $\log(NM)$).
\end{proof}
\subsection{\textbf{The Proof of Corollary III.1}}
\begin{proof}
	Based on the relation between the 2D histogram and marginal histogram the following inequality holds.
	\eas{
		&p_i(\Cv) \geq p_{i,j}(\Cv) \ \ \forall i,j \in [2^D-1]\\
		&\Leftrightarrow \log(\tilde{h}_i(\Cv)) \geq \log(\tilde{h}_{i,j}(\Cv))
		\label{cor1} \\
		&\Rightarrow 
		-\sum_{ \{j,i\} \in [2^D-1]}\tilde{h}_{i,j}(\Cv)\dfrac{\log(p_{i}(\Cv))}{\log(\tilde{h}_{i,j}(\Cv))}
		\\ & \geq -\sum_{ \{j,i\} \in [2^D-1]}\tilde{h}_{i,j}(\Cv)\dfrac{\log(p_{i}(\Cv))}{\log(\tilde{h}_{i}(\Cv))}\\
		& =-\sum_{i  \in [2^D-1]}\dfrac{\log(p_{i}(\Cv))}{\log(\tilde{h}_{i}(\Cv))}\sum_{j  \in [2^D-1]}\tilde{h}_{i,j}(\Cv)\\
		&=-\sum_{i  \in [2^D-1]}\tilde{h}_i(\Cv)\dfrac{\log(p_{i}(\Cv))}{\log(\tilde{h}_{i}(\Cv))}=r(\Cv,\Xv_1).
	}
	By analogy with this method, the same inequality can be obtained for the second image.
\end{proof}

\end{document}